\documentclass{article}
\usepackage{spconf,amsmath,graphicx}
\usepackage{algorithm}
\usepackage{algorithmic}
\usepackage{listings}
\usepackage{hyperref}
\title{ENERGY CONSUMPTION OF DEEP GENERATIVE AUDIO MODELS}
\name{Constance Douwes$^{\star}$ \qquad Philippe Esling$^{\star}$ \qquad Jean-Pierre Briot$^{\dagger \ddagger}$}
  
  \address{$^{\star}$IRCAM, Sorbonne Université, CNRS, UMR 9912 F-75004 Paris, France\\
      $^{\dagger}$Sorbonne Université, CNRS, LIP6, F-75005 Paris, France\\
	$^{\ddagger}$UNIRIO, Rio de Janeiro, RJ 22290-250, Brazil}

\begin{document}

\maketitle

\begin{abstract}
In most scientific domains, the deep learning community has largely focused on the quality of deep generative models, resulting in highly accurate and successful solutions. However, this race for quality comes at a tremendous computational cost, which incurs vast energy consumption and greenhouse gas emissions. At the heart of this problem are the measures that we use as a scientific community to evaluate our work.

In this paper, we suggest relying on a multi-objective measure based on Pareto optimality, which takes into account both the quality of the model and its energy consumption. By applying our measure on the current state-of-the-art in generative audio models, we show that it can drastically change the significance of the results. We believe that this type of metric can be widely used by the community to evaluate their work, while putting computational cost -- and \textit{in fine} energy consumption -- in the spotlight of deep learning research.
\end{abstract}


\section{Introduction}
\label{sec:intro}

Most of the recent advances produced by deep approaches rely on a significant increase in terms of both size and complexity \cite{Hernandez2020}, as well as an ever-growing number of training examples. Hence, such improvements are often only permitted by a concomitant increase in power consumption \cite{Thompson2020} and, thus, carbon emissions \cite{Strubell2020a}. In the audio synthesis domain, deep generative models have reached an unprecedented quality for waveform synthesis (in both speech and music). Researchers usually concentrate on high-quality synthesis, which is a major challenge given that raw waveforms require handling intricate temporal structures at both local and global scales. Therefore, models are usually highly complex and computationally expensive, with either extremely large recurrent cells or convolutions. The training time needed for them to converge along with their complexity questions their real effectiveness with regards to the quality of the generated results.

\begin{figure}[!ht]
    \centering
    \includegraphics[width=\columnwidth]{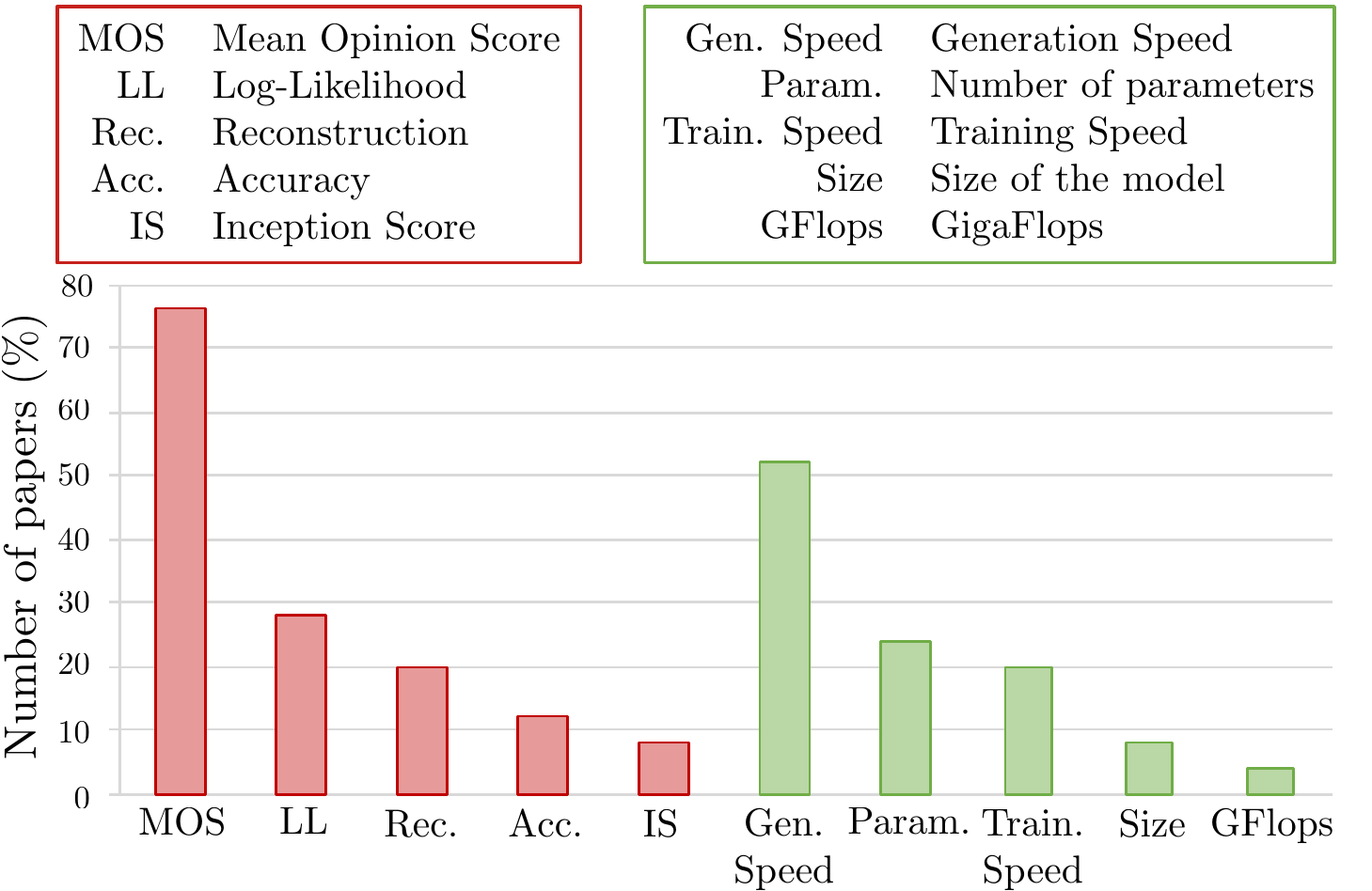}
    \caption{Distribution of commonly-used measures to compare and evaluate  generative audio models. In red (left) those that refer to the quality of the generated samples, and in green (right) those that refer to their algorithmic complexity and performances.}
\label{figure:distribution}
\end{figure}

Generally speaking, the absence of energy consumption criteria for generative models falls within the broader problem of lacking evaluation methods, notably for assessing their quality \cite{Theis2016}. Figure~\ref{figure:distribution} displays the distribution of different evaluation metrics used in twenty-five state-of-the-art neural audio synthesis research papers. As we can see, current researches focus on generation quality rather than on computational performances when evaluating and comparing models. Although some studies do mention the training time per iteration and the number of generated samples per second, energy consumption has never been taken into account, neither for training nor for inferring new sample. However, measuring the precise energy consumption of a given model is a complex endeavor \cite{Garcia-Martin2019}, which remains mostly neglected. The recent domain of \textit{green computing} aims to address this kind of issues \cite{Schwartz2019}. Although this is a novel field in deep learning research, it is already emerging in some communities such as Natural Language Processing (NLP) \cite{Strubell2020}.

In this work, we propose a new method to evaluate the energy efficiency of any generative models, while analyzing specifically the task of raw waveform generation. First, we introduce estimations of the training costs for all state-of-the-art models for which we had enough training details among the twenty-five used in Figure \ref{figure:distribution}:  \textit{SampleRNN} \cite{Mehri2019}, \textit{SING} \cite{Defossez2018}, \textit{WaveGAN} \cite{Donahue2019}, \textit{GANSynth} \cite{Engel2019} and \textit{FloWaveNet} \cite{Kim2019a}. Then, we propose the use of a multi-objective Pareto optimality criterion to provide fair comparisons simultaneously on generation quality and energy efficiency. For that purpuse, we focus on a recent model called \textit{WaveFlow} \cite{Ping2019}, and measure the effective energy consumption of training and inferring new samples for five alternative configurations.

\section{Generative audio models}

Generative models are a flourishing class of machine learning approaches, which deal with learning to generate novel data based on the observation of existing examples. Given training data points $\mathbf{x}$ following an unknown probability distribution $p(\mathbf{x})$, generative models aim to learn a parametric distribution $p_{\theta}(\mathbf{x})$ that approximates $p(\mathbf{x})$, by iteratively changing model parameters $\theta$. Several methods exist to address this task, which can be split in four categories: \textit{auto-regressive models}, \textit{Variational Auto-Encoders} (VAE) \cite{Kingma2014}, \textit{Generative Adversarial Networks} (GAN) \cite{Goodfellow2014} and \textit{Flow} models \cite{Rezende2015}. 

Auto-regressive methods attempt to model examples by assuming that the output variable is linearly related to its prior value. Following this, \textit{WaveNet} \cite{Oord2016} and \textit{SampleRNN} \cite{Mehri2019} have tackled direct waveform learning and generation. Unfortunately, these methods are based on heavy architectures whose computational complexity incur humongous energy consumption, both for training and inference. Furthermore, they also provide almost no direct control on the generative process. Some approaches use \textit{VAEs} \cite{Engel2017} that learn a latent space providing a low-dimensional representation of the data while remaining rather simple and fast to train. However, the generated samples tend to be slightly blurry compared to recent adversarial networks, such as \textit{WaveGan} \cite{Donahue2019} or \textit{GANSynth} \cite{Engel2019}. These show impressive reconstruction abilities but lack latent expressivity and are difficult to optimize due to unstable training dynamics. The recently proposed \textit{Normalizing Flows} (NF), used in \textit{WaveFlow} \cite{Ping2019} or \textit{FloWaveNet} \cite{Kim2019a} allow to model highly complex distributions in the latent space and already yield remarkable results.
%add the fact that there are no data space reduction? 

\section{Energy efficiency measures}

In deep learning contexts, we consider two types of energy consumption: the amount of energy required to \textit{train} a model (until convergence), and the amount of energy required by the model for \textit{inference} (generating a sample in the case of audio synthesis). Measuring the energy consumption of any kind of computer program is already a challenging task, since there are many variables involved, e.g. cache hits, cache misses, DRAM accesses. A popular metric is the kiloWatt-hour (kWh). As its name suggests, it is the multiplication of the power in kilo-Watts by the time in hours. The traditional way to get the instantaneous power is to use a Watt-meter to manually measure it on the hardware. The main drawback of this technique is that deep models mostly work with GPUs on distant servers. Fortunately, Lacoste et al. \cite{Lacoste2019} proposed a simpler online tool called the "Machine Learning Impact Calculator"\footnote{\url{https://mlco2.github.io/impact/}} that provides an approximation of the carbon emission required for training a model, by considering the location of the servers, the total training time, and the hardware on which the training takes place. Very recently, Anthony Wolff et al. \cite{WolffAnthony2020} developed an open-source tool written in Python called "Carbontracker"\footnote{\url{https://github.com/lfwa/carbontracker}}, which tracks and predicts energy consumption and carbon emissions for training deep leaning models. This provides a more accurate estimation while being user-friendly.

Another common measure of energy efficiency is the total number of model parameters as it is quite easy to determine and usually directly correlated with computational complexity. Unlike aforementioned measures, this one is hardware- and location- independent. Nonetheless, the number of parameters does not accurately reflect power consumption as some operations consume more than others. Hence, the best way to alleviate that issue is to consider the number of Floating Point Operations (FPO) of a model \cite{Schwartz2019}. This computation is not straightforward : it depends on each layer’s characteristics, e.g., input size, kernel size, stride, padding, bias. A Python package exists called “PyTorch-OpCounter”\footnote{\url{https://github.com/Lyken17/pytorch-OpCounter}}, but researchers have to include their calculations when using different types of layers than the implemented ones. 

\section{Estimations of energy consumption for existing models}
\label{sec:estimations}
\subsection{Models}
\label{sec:models}
After a review of the state-of-the-art neural audio synthesis models working directly on waveform \cite{Tan2021,Huzaifah2021}, we selected those for which we had enough training details to estimate their energy consumption. These include the hardware used to train the model, such as the type of GPU and total training time in hours. Surprisingly, we found out that only five of the studies properly specified both criteria. Here, we present a short description of these models and the details of their training procedure according to the original papers. 

\textit{SampleRNN} introduced by \cite{Mehri2019} is an auto-regressive model predicting one sample at a time. It is composed of hierarchical recurrent layers working at different temporalities. This model is trained for about one week on a GeForce TITAN X on three different datasets containing speech, vocal sounds and piano sonatas.

\textit{SING} proposed by \cite{Defossez2018} is a convolutional  audio synthesizer that generates waveform given desired categorical inputs. The training is composed of three parts: first, an auto-encoder is trained for 12 hours, then a sequence generator for 10 hours and finally an end-to-end fine-tuning for 30 hours. All parts are trained on 4 NVIDIA P100 GPU on the \textit{NSynth} dataset.

\textit{WaveGAN} \cite{Donahue2019} is a GAN that performs raw waveform synthesis using transposed convolutions acting as upsampling modules. The network is trained on a single NVIDIA P100 GPU and converges within 4 days on the \textit{Speech Commands Zero Through Nine} dataset.

\textit{GANSynth} \cite{Engel2019} also uses GANs to generate log-magnitude spectrograms and phases instead of modeling the raw waveform directly. The training lasts for 4.5 days on a NVIDIA V100 GPU on a subset of the \textit{NSynth} dataset. 

\textit{FloWaveNet} proposed by \cite{Kim2019a} is a flow-based model for parallel waveform speech synthesis using the WaveNet architecture as an inverse transformation function. The training requires 11.3 days on a NVIDIA Tesla V100 GPU and operates on the \textit{LJSpeech} dataset.

Although we selected these five models for the availability of their training details and not for their specific architecture, we believe that they form a representative set of the current generative models, even though there are no VAE-based models.

\subsection{Training cost estimations}
\label{sec:energycost}
Since we do not possess all of the previously mentioned hardware, some hypotheses have to be taken into account. As Machine Learning Impact Calculator does, we measure the  kilo-Watt hours consumption for the worst-case scenario. To do so, we take the maximum power of consumption  for each GPU according to their technical specifications and multiply it by the number of GPUs used for training and then by training time in hours. Note that the percentage of GPU utilization is not equivalent to the percentage of power consumption.

\begin{table}[h]
\begin{center}
\begin{tabular}{llrrr}
Model & Hardware & Power & Hours & Energy \\
\hline
\hline
FloWaveNet  & 1 $ \times$ V100 & 300 & 272 & 81.6 \\
GANSynth    & 1 $ \times$ V100 & 300 & 108 & 32.4 \\
SampleRNN  & 1 $ \times$ TITAN X & 250 & 168 & 42 \\
SING       & 4 $ \times$ P100& 250 & 52 & 52 \\
WaveGAN   & 1 $ \times$ P100 & 250  & 96 & 24 \\
\end{tabular}
\caption{Approximated energy consumption for training several state-of-art generative audio models. Power is expressed in Watts and energy in kWh.}
\label{table:table1}
\end{center}
\end{table}

Estimations are shown in Table \ref{table:table1}. As a reference point, the daily average electricity consumption is 12.1 kilowatt hours (kWh) per person in the U.S. This shows that the energy of a training procedure is far from marginal and should be more widely discussed. Furthermore, these estimations do not include the total amount of energy used to train and test all the different versions to find the right architectures and configurations. 

To see if our predictions are close to real energy demand, we measure the consumption of training SING (auto-encoder, sequence generator, and fine-tuning) on a single TITAN X with the same configuration as the original paper. We use Carbontracker \cite{WolffAnthony2020} to predict the actual energy consumption of the whole learning process. We found out that, on our hardware, it consumes 64.8 kWh, which is slightly higher than our initial estimation (certainly due to CPU and DRAM energy draw). At this point, we believe that comparing models purely on the basis of these estimations is questionable, and argue that the real energy should have been recorded. Besides, although heavy models trained on large datasets may consume more, they are often more accurate, but is this gain worth the additional energy consumption? As a result, we advocate using a multi-criteria evaluation to find the best compromise between quality and energy consumption, which is the purpose of the next section.

%This table normally belong to part 5, but thanks to this fabulous latex compiler I have to put it here to have a good render. 
\begin{table*}[h]
\begin{center}
\begin{tabular}{lrrrr}
Model & \# param & 1-\%MOS & E$_{train}$ (kWh) & E$_{gen}$ (Wh)\\
\hline
\hline
WaveFlow 1 ($h=8$, $r=64$) & 5.91 M & 0.148 & 407.7 & 1.349\\ 
WaveFlow 2 ($h=16$, $r=64$)& 5.91 M & 0.136 & 437.6 & 1.382\\
WaveFlow 3 ($h=16$, $r=96$)& 12.78 M & 0.132 & 725.4 & 2.382\\
WaveFlow 4 ($h=16$, $r=128$)& 22.25 M & 0.124 & 644.8 & 2.512\\
WaveFlow 5 ($h=16$, $r=256$)& 86.18 M & 0.114 & 1011.2 & 3.871\\
\end{tabular}
\caption{Subjective score ($1-\%\text{MOS}$) for multiple configuration deduced from \cite{Ping2019} and their number of parameters. E$_{train}$ and E$_{gen}$ stands respectively for the amount of energy required for a whole training, and the amount of energy to produce 100 seconds of raw audio at 22.05 kHz. $h$ is the squeezed height and $r$ the residual channels, for more information see \cite{Ping2019}.}
\label{table:waveflow}
\end{center}
\end{table*}

\begin{figure*}[h]
    \centering
    \includegraphics[width=\textwidth]{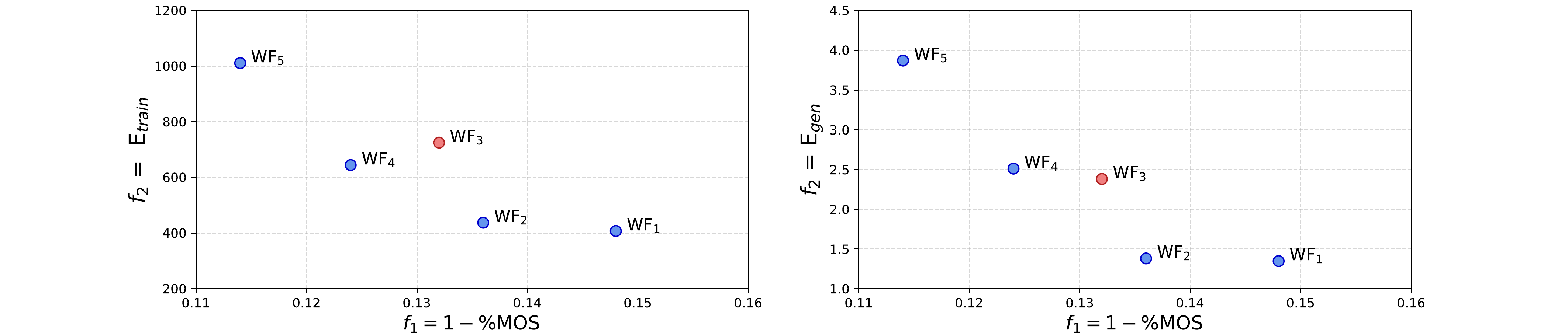}
    \caption{Representation of two Pareto space for optimizing quality $f_1$ and energy $f_2$ of either the training cost (left) or the inference cost (right) of generative audio models, where WF$_i$ stands for WaveFlow $i$. In blue, all optimal solutions, while in red dominated ones.}
    \label{figure:paretoWF}
\end{figure*}

\section{MULTI-CRITERIA EVALUATION : USE CASE}
\label{sec:multicriteria}

\subsection{Methodology}

First, we present the quality score we use for the study, and then the training and the inference cost. As discussed earlier, measuring the quality of generative models is a daunting task. The plurality of metrics used in the literature (Figure \ref{figure:distribution}) comes with the diversity of architectures proposed. Indeed, no straightforward objective score can be computed in generative tasks (apart from reconstruction rate in the case of AE-based generative models), as opposed to classification or prediction tasks . The most popular and relevant measure across the audio generation literature is the Mean Opinion Score (MOS). It is a subjective measure ranging from 1 to 5, based on a qualitative test where participants are asked to rate as 1 the lowest perceived quality and 5 the highest when comparing a set of results. For our purpose, we compute $\%\text{MOS}=\text{MOS}\div5$ to re-scale this measure between 0 and 1 and take $1-\%MOS$ as our first minimization task. The higher the perceived quality of the sound, the closer this measure will be to 0, and conversely the lower the perceived quality, the closer it will be to 1. 

Unfortunately, only three out of the five models from Section \ref{sec:models} use the MOS, and no cross-comparisons were conducted. Since this measure is highly dependent of the dataset on which the model was trained and the experimental setup, it is not meaningful to compare MOS from different papers to one another. Therefore, we conduct the rest of the experiments on a model that provides a large number of MOS for a wide range of configurations while remaining relatively efficient: WaveFlow \cite{Ping2019}. We rely on a PyTorch implementation\footnote{\url{https://github.com/L0SG/WaveFlow}}, and use the same configurations as the original paper. 
\newpage
To measure the energy consumption of each of the learning procedure, we train models on 4 TITAN V with a batch size of 8. As previously, we use Carbontracker to predict the global training cost. We compute the number of iterations per epoch and derive the number of epoch needed since the original paper does not specify the number of epoch  but rather the number of steps: lighter models (res. channels 64 and 96) are trained for 3M steps, medium ones (res. channels 128 and 96) for 2M and the heaviest one (res. channels 256) for 1M due to time constraints. Unlike training, which occurs only once, a model can infer an unbounded number of times. Thus, we choose to conduct experiments on the energy required to produce and record 10 audio clips of 10 seconds at 22.05kHz on a single TITAN V GPU. 

\subsection{Results}

We summarize in Table \ref{table:waveflow} the subjective measure $1-\%MOS$ deduced from the original paper \cite{Ping2019} along with the number of parameters for each of the five configurations studied. We also include our measurement of the energy required to train E$_{train}$ in kWh and generate E$_{gen}$ in Wh. As we can see, increasing the size of a model and the amount of training examples increases its quality, but also the energetic cost. Our idea is to rely on Pareto optimization, which is is a branch of mathematical optimization problems involving several conflicting objectives to be optimized simultaneously. This notion is used when it is impossible to improve one objective without degrading another. We display in Figure \ref{figure:paretoWF} two multi-objective space. The one on the left accounts for training efficiency, while the one on the right accounts for generation efficiency. Models that are Pareto optimal are shown in blue, while those that are not are shown in red. As we can see, the multi-objective measure shows that model $\text{WF}_{3}$ is Pareto-dominated by others. Hence, this underlines the interest of conducting a joint analysis on both criteria simultaneously.

\section{CONCLUSIONS}
\label{sec:conclu}

According to our knowledge, this is the first study on energy consumption for waveform generation and a primer attempt to include energy efficiency in the entire evaluation procedure. First, we used indications (hardware and training time) from existing generative audio models papers to approximate energy consumption of training procedures. While increasing awareness, we also showed that this calculation must be linked to the quality of the models. To that end, we proposed the use of a new evaluation based on Pareto optimality to give an equivalent importance to both model quality and their energy consumption. This places computational complexity and resources at the heart of the research process. It should be noted that our approach is generic and could be applied to any type of model or input data.

%I may change references tomorrow cause I just saw that everything is a preprint ... 
\vfill\pagebreak

\bibliographystyle{IEEEbib}
\bibliography{library}

\end{document}